# Natural Language Processing and Deep Learning Models to Classify Phase of Flight in Aviation Safety Occurrences

Aziida Nanyonga
School of Engineering and Information Technology
University of New South Wales
Canberra, Australia
a.nanyonga@adfa.edu.au

Oleksandra Molloy
School of Engineering and Information Technology
University of New South Wales
Canberra, Australia
o.molloy@unsw.edu.au

Ugur Turhan
School of Engineering and Information Technology
University of New South Wales
Canberra, Australia
u.turhan@adfa.edu.au

Hassan Wasswa
School of Engineering and Information Technology
University of New South Wales
Canberra, Australia
h.wasswa@adfa.edu.au

Graham Wild
School of Engineering and Information Technology
University of New South Wales
Canberra, Australia
g.wild@adfa.edu.au

*Abstract*— The air transport system recognizes the criticality of safety, as even minor anomalies can have severe consequences. Reporting accidents and incidents play a vital role in identifying their causes and proposing safety recommendations. However, the narratives describing pre-accident events are presented in unstructured text that is not easily understood by computer systems. Classifying and categorizing safety occurrences based on these narratives can support informed decision-making by aviation industry stakeholders. In this study, researchers applied natural language processing (NLP) and artificial intelligence (AI) models to process text narratives to classify the flight phases of safety occurrences. The classification performance of two deep learning models, ResNet and sRNN was evaluated, using an initial dataset of 27,000 safety occurrence reports from the NTSB. The results demonstrated good performance, with both models achieving an accuracy exceeding 68%, well above the random guess rate of 14% for a seven-class classification problem. The models also exhibited high precision, recall, and F1 scores. The sRNN model greatly outperformed the simplified ResNet model architecture used in this study. These findings indicate that NLP and deep learning models can infer the flight phase from raw text narratives, enabling effective analysis of safety occurrences.

## I. Introduction

Air transportation is a highly safety-critical industry that is susceptible to operational errors, making it one of the most sensitive fields in the transport sector. Even the smallest mistake or minor system misconfiguration can lead to catastrophic disasters, resulting in various consequences such as loss of lives, significant financial losses, erosion of customer trust, and damage to property, among other impacts [1]. In the event of an aviation safety incident, regardless of its severity, reporting, and investigations are essential.

The primary objective of these safety reports is not to assign blame but rather to ensure that similar incidents are prevented from occurring again. Aviation safety reports, which encompass both formal investigations and voluntary self-reports, consist of unstructured text narratives presented in a language understandable to humans [2-5]. These narratives highlight the sequence of events that are potentially contributing or causal factors of the incident or accident. Moreover, these reports are made publicly available in some jurisdictions. The publication of these reports serves multiple purposes, including facilitating access for various stakeholders within the aviation industry. This includes maintenance teams, operational managers, and researchers in the field of aviation safety, who can utilize the reports to access, analyze, critique, and implement or adhere to the recommended safety standard operating procedures, rules, and regulations.

However, when similar incidents occur, investigation teams often need to refer to previous reports, which can be a time-consuming process, especially if the conditions of the current incident are related to multiple past incidents. The challenge arises because, although most reports provide a tabular summary of the conditions surrounding the event to enhance readability for humans, the detailed sequence of events that likely caused the incident or accident is typically described in narrative form using natural language understood by humans. Furthermore, when humans report such narratives, they often use non-standard terms that are comprehensible to other humans but not to computers. Consequently, if there is a need to revisit these reports, humans are required to search through a large database of reports and read them one by one in a sequential manner. This manual approach significantly delays the investigation process. To overcome this challenge, various models have been proposed to expedite the processing of aviation data and assist aviation authorities and stakeholders in making safety-critical decisions. Typical tools of decision making, and risk assessment (operations research) can be utilized for modelling and forecasting in aviation safety, such as multi-attribute utility modelling [6]. More fundamental statistical or probability methods can be utilized as well, such as Bayesian networks [7]. The resultant models enable faster data analysis and decision-making, providing valuable insights to support the industry.

Furthermore, advanced machine learning algorithms such as Artificial Neural Networks (ANN) [2, 8], Random Forest (RF) [9], and Support Vector Machine (SVM) [10, 11], among others, have been proposed as state-of-the-art models for predicting similar outcomes. However, most of these studies have primarily focused on structured data, neglecting the unstructured text narrative. Some studies have used models like Latent Dirichlet Allocation (LDA) [12] in natural language processing applications for aviation incident reports, primarily for topic modeling tasks [12, 13]. Although topic modeling can classify safety data reports into different topics, its effectiveness diminishes as the overlap and ambiguity between topics increase, as many words can be assigned to multiple topics, rendering it highly unreliable. Despite the increasing popularity and adoption of deep learning models in various domains, including health and NLP for spam email detection [9, 14], the aviation industry has not fully embraced deep learning models for big data analysis.

In this study, an approach that utilizes natural language processing (NLP) and artificial intelligence (AI) models is presented to process text narratives and classify the flight phases of safety occurrences. While previous studies have explored NLP techniques for text analysis in aviation safety, this research extends the use of deep learning models, namely ResNet and sRNN, to aviation safety for classifying the Flight phase at which the incident occurred. This application of deep learning algorithms to infer flight phases from raw text narratives in safety occurrences represents a unique contribution to the field.

The motivation behind this study lies in the potential benefits it offers to the aviation industry. By accurately classifying safety occurrences into specific flight phases, such as take-off, landing, approach, enroute, maneuvering, and more, we can identify stressors or patterns that may lead to component failures. This proactive identification allows airlines and maintenance crews to address issues and reduce unplanned maintenance events. Furthermore, our approach enables aviation safety professionals to identify potential risks and hazards associated with different phases of flight, particularly during the safety-critical phases at the beginning and end of an operation (take-off through initial climb, and final approach to landing).

Therefore, the main objective is to compare the performance of ResNet and sRNN algorithms in classifying the flight phase during which the safety occurrence event took place. The classification can utilize supervised learning for the flight phase categories, as this is provided in the NTSB aviation accident database. Specifically, aviation safety occurrences (accidents and incidents) are classified with the relevant phase of flight, such as Landing, Approach, Take-off, Enroute, Maneuvering, etc. This enables airlines and maintenance crews to proactively address issues and reduce unplanned maintenance events.

The rest of this work is organized as follows. Section III gives an account of the existing literature regarding aviation safety and machine learning. Section IV gives details of the proposed approach and the implementation procedure followed to realize it. In section V the results of the study are presented and give a detailed discussion and their implication for aviation safety research. Section VI presents the conclusion highlighting the direction of future work.

## II. RELATED WORK

Machine learning has been widely applied in various fields, including aviation, medicine, bioinformatics, and biology [17]. In the context of aviation safety reports, machine-learning techniques have been utilized for data mining and text classification. Recent studies have explored different approaches for text mining and classification in aviation safety reports.

Another study [18] examined the application of various recurrent neural network (RNN) architectures for sentence modelling, showing that these models are suitable for sequential datasets such as text mining. Another study [19] used RNN techniques to predict weather-related tasks and regulate pre-flight information. Similarly, a study by Paul [20] focused on different natural language processing (NLP) techniques applied to civil aviation corpus, suggesting the use of NLP, particularly RNN, for time series data mining.

In a study analyzing aviation safety reports, Chanen [21] proposed a deep learning approach using word2vec models to identify similar terms and semantic relationships within the reports. The model aimed to improve the understanding of reporting experts and reduce uncertainty in safety reports. However, the study only focused on one NLP tool, leaving room for improvement using different NLP approaches.

Another related study employed sequential deep learning [22], specifically LSTM and word embedding, for aviation safety prognosis using NTSB reports. Their classification models showed promise in reviewing and analyzing safety investigation reports. Similarly, ElSaid et al. [23] used LSTM models to predict aircraft engine vibrations, demonstrating that LSTM recurrent neural networks can accurately predict vibrations in flight datasets.

In the evaluation of ML techniques for determining critical causes of accidents in air transport [24], it was found that the decision tree algorithm performed better in predicting the cause of accidents. Their study highlighted the human factor, particularly loss of attention, as a significant cause of accidents in aviation.

A study from CLLE-ERSS research lab and CFH safety company [4], explored NLP tools and text-mining methods for analyzing unstructured data in aviation reports. The study emphasized the importance of NLP techniques in accident report analysis and compared various datasets from different databases. However, specific NLP tools used were not reported, indicating a potential for better performance with different tools.

In 2022, Ya et al. conducted a study on the application of deep learning, particularly ResNet, for radio signal recognition in real-world scenarios [25]. Their comprehensive research showcased the efficacy of deep learning models in accurately identifying and categorizing radio signals. Notably, ResNet demonstrated exceptional performance in managing extensive datasets and achieving substantial enhancements in recognition accuracy. These findings contribute to the progress of deep learning techniques for radio signal recognition, with potential applications across different fields, including the aviation industry. The study emphasizes the significance of this technology in advancing signal recognition capabilities.

In the context of F-16 fighter jet data modeling, Mersha et al. [15] tested different RNN architectures and found that sRNN using the trained estimated algorithm achieved the best performance in predicting certain parameters. Similarly, in analyzing railroad accident narratives, Song et al. [16] used

BLSTM, CNN, and RNN models to improve text classification accuracy, although their performance was below 80%.

A proposed approach [26] constructed a network based on Bayesian probability distribution to infer causal relationships between aviation accidents. However, the study used an outdated dataset that does not capture current trends in aviation data. Another study [27] proposed a data management scheme using ontologies and conceptual models to enhance aviation safety data management, but it lacked advanced data analytics and machine learning techniques.

Lastly, [28] utilized principal component analysis (PCA) and deep brief networks for flight incident prediction. Although the deep brief network architecture showed benefits in learning complex features, PCA alone may lead to reduced classification performance, potentially compromising aviation safety.

Overall, these studies demonstrate the application of machine learning and NLP techniques in analyzing aviation safety reports and improving safety measures. However, there is still room for further exploration and improvement in utilizing advanced data analytics and adapting to the dynamic nature of the aviation industry.

### III. METHODOLOGY

To implement the proposed approach, the research involved several distinct steps, including the identification of the training dataset, text processing, and classification, as depicted in Figure 1. A brief description of the utilized datasets is provided in the following subsection.

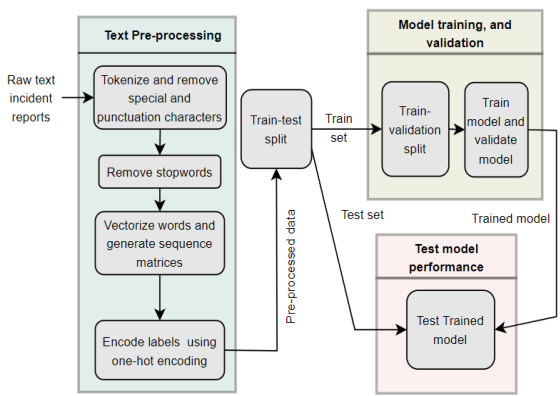

Fig. 1. Methodological framework

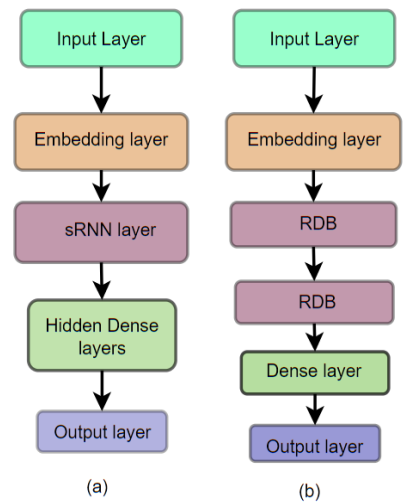

Fig. 2. Deep learning architectures: (a) simple RNN and (b) ResNet with two residual dense blocks

### A. Data Acquisition

Aviation incident/accident investigation reports are collected and published by various organizations such as ATSB (Australian Transport Safety Bureau), ASRS, and NTSB. For this study, the researchers utilized the NTSB aviation incident/accident investigation reports[1]. Depending on the specific problem being addressed, this dataset can be obtained from the NTSB website, which offers different forms of metadata including individual monthly published PDF reports, and JSON files, or by querying individual reports online or through a CSV summary. In this research, researchers downloaded a JSON file containing incident/accident investigation details spanning from 2005 to 2020. Moreover, the researcher focused on incidents where investigations were completed, resulting in a dataset comprising 16,919 records following data preprocessing and cleaning. From each report, the 'analysis Narrative' and 'flight phase' fields were extracted for training and validating our deep learning models.

### B. Text Processing

Machine learning models are not inherently capable of understanding text data in its raw form. Therefore, the text data needs to be converted into a numeric format that can be interpreted by the models. In this study, the Karas deep learning library was employed due to its extensive collection of deep learning models and different types of model layers. This library also offers advanced modules for text pre-processing. The Tokenizer module, for instance, generates tokens and sequence vectors for the input text. The categorical module is used to map unique categorical entries of the damage level variable (e.g., destroyed, substantial, minor, and none) to corresponding numerical values or as one-hot encoded entries for each data instance.

To handle unwanted special and punctuation characters, remove stop-words, and perform word lemmatization, the spacy library was utilized. Spacy is a Python library specifically designed for text-processing tasks, including named entity recognition and word tagging. It incorporates a comprehensive list of special characters, punctuation, and stop-words, and is regularly updated when necessary.

---

[1] https://data.ntsb.gov/carol-main-public/query-builder?month=1&year=2020

With the a fore mentioned tools, each input text narrative was processed and transformed into a representative sequence or vector with a length of 2000. Numeric sequences derived from text narratives with fewer than 2000 words were padded with zeros, while those exceeding 2000 words were truncated. The vocabulary size of the corpus was set to 100,000.

To split the dataset into training, validation, and testing sets, the train-test-split module from scikit-learn was employed. All experiments in this study were conducted using Python as the programming language, with Jupyter Notebook serving as the code editor.

*C. Text Classification*

The data set is pre-processed and divided randomly into an 80% training set and a 20% test set. Furthermore, during the training process, 10% of the training set is set aside for model validation in each epoch. Two deep learning models, ResNet and sRNN, are trained using this data, and their performance is assessed. The Adam optimizer is employed for model optimization. However, it is worth noting that this study did not focus on identifying the best optimizer, so alternative optimization techniques can be utilized if desired.

*D. Deep learning model Architecture*

To ensure consistency, a shared architecture was utilized for all models, with minor variations in the case of combined models. This common architecture consisted of an embedding layer, hidden layers, and an output layer. ReLU (Rectified Linear Unit) activation function was employed for all hidden layers, while the SoftMax activation function was used for the output layer. The predicted class was generated using the argmax function, which identifies the index associated with the highest probability in the SoftMax output. Figure 2 provides an overview of the deep learning architectures employed in this research.

*1) ResNet Architecture:*

The ResNet architecture consists of several layers organized into blocks. Each block contains multiple residual units, which are the basic building blocks of ResNet. In a ResNet, the input passes through several blocks, and each block contains a set of residual units. The output of each block is fed into the next block until the final output is obtained. The basic residual unit comprises two or three convolutional layers, followed by element-wise addition with the input. The key concept behind ResNet is residual learning, which involves learning the residual mapping instead of learning the entire mapping from input to output directly. This is achieved by introducing skip connections or shortcut connections that allow the network to learn the residual mapping more effectively [29].

In this work, a simplified ResNet model was built constituting only two residual dense blocks (RDB) as shown in Figure 2 above, and with each RDB having only 3 skip connections as shown in Figure 3. The skip connection enables the gradient to flow directly from the later layers to the earlier layers during backpropagation. This alleviates the vanishing gradient problem and facilitates the training of deeper networks. Additionally, it helps to preserve the learned information from earlier layers, making it easier for the network to capture fine-grained details and avoid degradation in performance as the network depth increases.

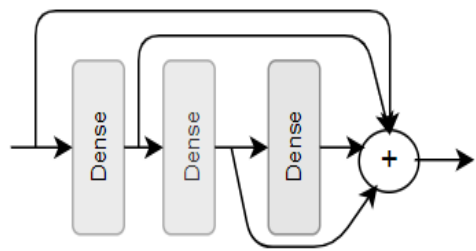

Fig. 3. Residual Dense Block

The formulas for the forward pass of each residual unit within the residual dense blocks can be summarized as follows:

Given an input x, the output y of a residual unit can be computed as:

$y = f(x, W) + x$,

where $f(x, W)$ represents the transformation performed by the convolutional layers in the residual unit with learnable weights W.

The "+" operator denotes the element-wise addition of the transformed input ($f(x, W)$) and the original input (x). This additionally combines the residual or the difference between the input and the desired output with the original input [30].

*2) Simple Recurrent Neural Networks (RNNs).*

Simple RNNs, introduced by Elman in 1990, are a fundamental type of RNN architecture. They consist of recurrent connections that allow information to flow in a loop, creating a feedback mechanism. This loop enables Simple RNNs to retain and utilize information from previous time steps, making them suitable for sequential data analysis.

Embedding layers are essential components in many sequence analysis tasks, such as natural language processing. They convert discrete inputs, such as words, into continuous, dense vector representations. Embeddings capture semantic relationships between inputs, enabling the network to better understand the underlying patterns in the data.

A simple Recurrent Neural Network (RNN) is an architecture that processes sequential data by maintaining a hidden state that captures information from previous inputs. It is a type of neural network designed to handle sequential or time-series data, where the order of the data points matters.

Hidden layers are the primary components responsible for learning and extracting meaningful representations from the sequential input. They consist of neurons that process the input at each time step and pass information to subsequent time steps. The depth and width of hidden layers determine the model's capacity to capture complex patterns and relationships in the sequential data.

The concept of a simple RNN can be explained using the following formulas:

Hidden state calculation:

$h_t = \text{activation}(W\_hh * h\_{t-1} + W\_xh * x\_t + b\_h)$

In this formula, h_t represents the hidden state at time step t, x_t is the input at time step t, and activation is an activation function like the hyperbolic tangent (tanh) or the rectified linear unit (ReLU). W_hh is the weight matrix connecting the previous hidden state to the current hidden state, W_xh is the weight matrix connecting the input to the hidden state, and b_h is the bias term for the hidden state.

y_t = activation (W_hy * h_t + b_y)

Here, y_t is the output at time step t, W_hy is the weight matrix connecting the hidden state to the output, and b_y is the bias term for the output.

During training, the RNN processes the input sequence one-time step at a time, updating the hidden state based on the previous hidden state and the current input. The output is then computed using the updated hidden state. To train an RNN, a loss function is defined, such as mean squared error (MSE) or cross-entropy, comparing the predicted output to the target output. The gradients are then calculated through backpropagation through time (BPTT), and optimization algorithms like gradient descent or Adam are used to update the weights and biases of the network [31]

## IV. RESULTS AND DISCUSSION

In this section, the performance of the two models is presented and compared with each other in terms of accuracy, precision, recall, and F1 measure. Fig. 3 illustrates a visual comparison of the validation accuracy achieved using the two models. It is clear from Fig. 3. that sRNN captures more patterns for discrimination between the various flight phases. Also, model performance was evaluated on the unseen test set samples and the results in Table 1 reveal that sRNN outperformed ResNet in terms of all performance evaluation metrics.

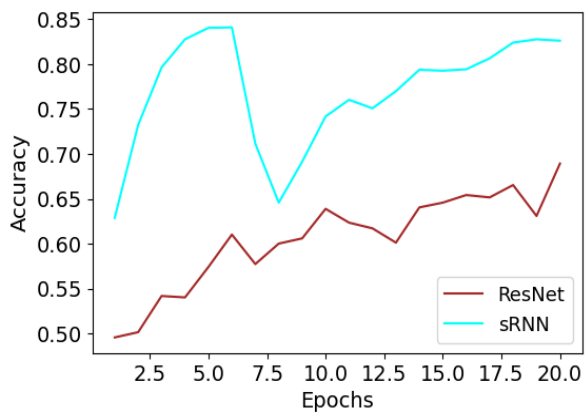

Fig. 4. Validation accuracy performance

TABLE I. DEEP LEARNING MODEL PERFORMANCE

| Models | Precision (%) | Recall (%) | F1 (%) | Accuracy (%) |
|---|---|---|---|---|
| **ResNet** | 69 | 69 | 68 | 68.9 |
| **sRNN** | 84 | 83 | 83 | 83.2 |

The findings of this study demonstrate that by employing natural language processing techniques in conjunction with advanced deep learning models, it is possible to predict the phase of flight at which an incident occurred based on the unstructured text narrative describing the events leading up to the accident.

## V. DISCUSSION

Our previous work focused on applying natural language processing (NLP) and deep learning models, such as LSTM, BLSTM, GRU, and sRNN, to classify and categorize safety occurrences in aviation based on text narratives. The objective was to determine the extent to which the damage level to an aircraft in a safety occurrence can be inferred from the text narrative using NLP techniques. The study evaluated the performance of various deep learning models individually and in combination, including sRNN. The models achieved competitive results, with an accuracy of over 87.9% and high precision, recall, and F1 scores.

In the current work, NLP is also applied to aviation incident/accident reports for text analysis, and advanced deep learning algorithms including ResNet and SimpleRNN are used for classification. The research question revolves around predicting the resultant flight phase based on the text narrative describing the series of safety occurrence events. The classification is based on the categories provided in the NTSB investigation reports, classifying the flight phase. The classification of the aircraft damage level and flight phase has significant implications for the decision-making process of the aircraft maintenance team. It helps determine whether the aircraft can be put back into operation, the extent of repairs required, and the associated budget, resource, and financial implications for the aviation company.

Overall, both the previous and current works demonstrate the utilization of NLP and deep learning techniques to analyze safety occurrences in the aviation industry using the NTSB dataset. While the previous work focuses on the classification of damage level in safety occurrences, the current work extends the analysis to aviation incident/accident reports with the aim of predicting the resultant flight phase based on textual narratives [32].

## VI. CONCLUSION

Safety is a fundamental aspect of the aviation industry and requires continuous monitoring and adherence to standard operating procedures, safety rules, regulations, and recommendations. Reports from investigations regarding aviation anomalies that pose a threat to stakeholders' safety often include textual narratives describing the sequence of events. In this study, different NLP techniques and deep learning models were utilized to classify the flight phase during which the incident occurred based on textual narratives of accidents/incidents from NTSB investigation reports. Through our preliminary experiments, the DL models achieved encouraging results, addressing the question of whether the flight phase can be inferred from the raw text narratives.

While deep learning models exhibit strong performance, they are often considered black-box models and further research should focus on enhancing interpretability and explainability of deep learning models used in aviation safety. This will foster stakeholder trust by enabling an understanding of the classification decisions and addressing concerns regarding the black-box nature of these models.